\documentclass[10pt,twocolumn,letterpaper]{article}

\usepackage{wacv}              

\usepackage{graphicx}
\usepackage{amsmath}
\usepackage{amssymb}
\usepackage{booktabs}

\usepackage{amsmath}
\usepackage{amssymb}
\usepackage{mathtools}
\usepackage{amsthm}
\usepackage{float}
\usepackage{amsmath,xspace,mathtools}
\usepackage{amssymb}
\usepackage{multirow}

\usepackage{enumitem}
\usepackage{algorithm}

\usepackage[accsupp]{axessibility}

\usepackage[pagebackref,breaklinks,colorlinks]{hyperref}

\newcommand{\mnist}{\textsc{MNIST}\xspace}
\newcommand{\cifar}{\textsc{Cifar-10}\xspace}

\newcommand{\ifc}{\operatorname{IFC}}
\newcommand{\ncd}{\textit{Non-Cross Diffusion}\xspace}
\newcommand{\xflow}{\textsc{xFlow}\xspace}

\usepackage[capitalize]{cleveref}
\crefname{section}{Sec.}{Secs.}
\Crefname{section}{Section}{Sections}
\Crefname{table}{Table}{Tables}
\crefname{table}{Tab.}{Tabs.}

\def\wacvPaperID{759} 
\def\confName{WACV}
\def\confYear{2025}

\begin{document}

\title{Non-Cross Diffusion for Semantic Consistency}


\author{Ziyang Zheng$^{*}$\quad
Ruiyuan Gao$^{*}$\quad
Qiang Xu
\\
The Chinese University of Hong Kong\\
{\small\texttt{\{zyzheng23,rygao,qxu\}@cse.cuhk.edu.hk}}
}

\twocolumn[{%
\renewcommand\twocolumn[1][]{#1}%
\maketitle
\begin{center}
\maketitle
  \centering
   \includegraphics[width=0.8\linewidth]{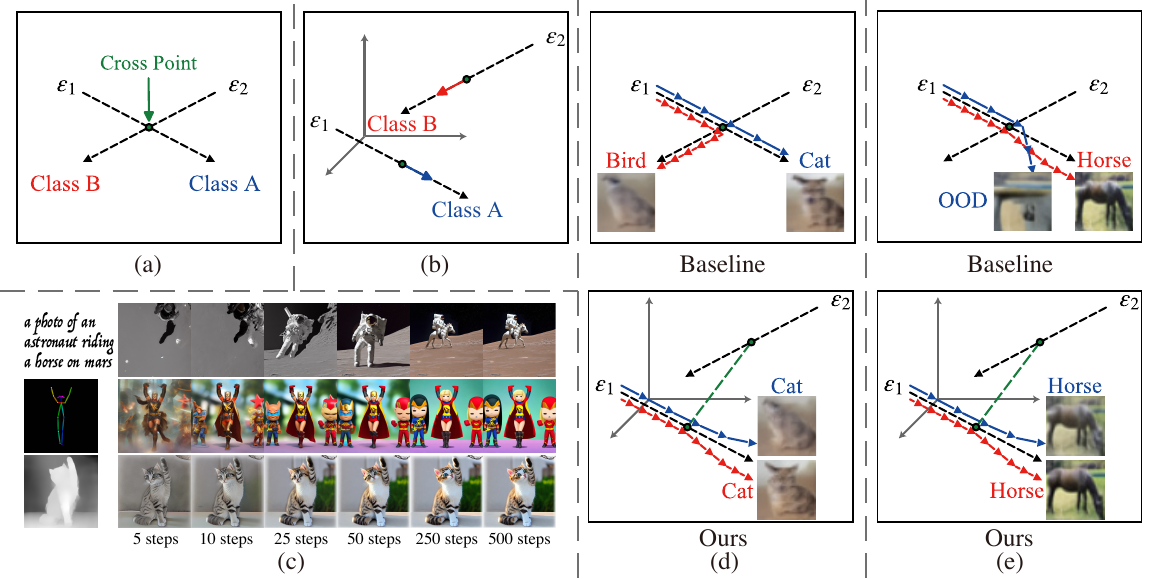}
   \vspace{-0.45cm}
    \captionsetup{type=figure, margin=1pt}
    \captionof{figure}{
  Illustrating \xflow in Diffusion Models. (a) Demonstrates the ambiguity in training targets caused by crossing flows, leading to the \xflow problem. (b) Shows how our method eliminates flow crossing by increasing the dimensionality of network inputs, thus resolving the \xflow problem.
 (c) Depicts how \xflow leads to variable sampling results across different steps, undermining deterministic sampling even for Stable Diffusion~\cite{LDM}.
 (d) \textit{Top}: Highlights the discrepancies between outcomes from reduced steps sampling (\textcolor{blue}{blue}) versus standard results (from 1000 steps \textcolor{red}{in red}) due to \xflow. \textit{Bottom}: Our method ensures consistent outputs across different sampling steps. (e) \textit{Top}: Exhibits instances where \xflow causes Out-Of-Distribution (OOD) outcomes in reduced steps sampling (\textcolor{blue}{blue}) compared to standard results (from 1000 steps \textcolor{red}{in red}). \textit{Bottom}: Our approach minimizes the occurrence of OOD samples.
    }
   \label{fig:OOD}
\end{center}
}]

\makeatletter
\def\blfootnote{\xdef\@thefnmark{}\@footnotetext}
\makeatother
\blfootnote{$^{*}$Equal contribution.}
\begin{abstract}

In diffusion models, deviations from a straight generative flow are a common issue, resulting in semantic inconsistencies and suboptimal generations. To address this challenge, we introduce `Non-Cross Diffusion', an innovative approach in generative modeling for learning ordinary differential equation (ODE) models. Our methodology strategically incorporates an ascending dimension of input to effectively connect points sampled from two distributions with uncrossed paths. This design is pivotal in ensuring enhanced semantic consistency throughout the inference process, which is especially critical for applications reliant on consistent generative flows, including various distillation methods and deterministic sampling, which are fundamental in image editing and interpolation tasks. 

Our empirical results demonstrate the effectiveness of Non-Cross Diffusion, showing a substantial reduction in semantic inconsistencies at different inference steps and a notable enhancement in the overall performance of diffusion models. 
\vspace{-10pt}
\end{abstract}

\section{Introduction}
\label{sec:intro}

\vspace{+0.15cm}

Diffusion models, as delineated in recent studies~\cite{DDIM,DDPM,diffusion_beat_gan,iDDPM,LDM,song2019generative,song2020score}, have exhibited remarkable capabilities in image synthesis, bolstering numerous applications such as text-to-image generation~\cite{glide,Imagen}, image editing~\cite{editing1,glide,editing2,editing3}, and image inpainting~\cite{editing1,inpainting1}. A key characteristic of these models is their multi-step generative process, which not only allows for correction of the diffusion path~\cite{song2020score} but also enhances controllability~\cite{shortcut,diffguard}.


Despite these advancements, the inference process in diffusion models typically involves a specific flow, whereas the training process entails random step selections from multiple flows. This randomness often results in a given training step correlating with diverse flows, creating ambiguity in target identification from the network's perspective, as depicted in Fig.~\ref{fig:OOD}(a). We term this phenomenon `\xflow'.

\xflow's emergence during training can hinder the model's optimization at certain steps, leading to a spectrum of generative issues. Notably, it challenges the model's ability to generate samples via a straight flow, compromising deterministic sampling across varying step counts, as shown in Fig.~\ref{fig:OOD}(c). It also complicates predicting later sampling steps from earlier ones, limiting the effectiveness of reward models~\cite {censored} and guided models~\cite{diffusion_beat_gan}. Moreover, in the context of distillation, which typically adopts a progressive approach, \xflow can introduce misleading signals, as evidenced in Rectified Flow~\cite{rectified_flow}. Perhaps most critically, \xflow can lead to the generation of Out-Of-Distribution (OOD) samples or low-quality samples, especially as sampling step size increases, as illustrated in Fig.~\ref{fig:OOD}(d-e).

In this paper, we propose a novel training strategy aimed at resolving the \xflow challenge in diffusion models. Our method centers on augmenting the input dimensionality to these models, a change that effectively prevents flow crossing. As depicted in Fig.~\ref{fig:OOD}(a), the issue at hand arises when two flows intersect, creating ambiguity; the input to the network (for instance, a noisy image) remains constant, yet it is associated with multiple potential targets (such as distinct noises originating from different images). To address this, our approach entails predicting the flow itself during the training phase, as shown in Fig.~\ref{fig:OOD}(b). Notably, we utilize the noise predicted by the network as the flow's endpoint, incorporating this element into the model's input. This technique sidesteps the pitfall of using groundtruth noise as input, which would otherwise result in trivial training solutions devoid of substantive learning. For practical implementation, we found ControlNet~\cite{ControlNet} particularly effective in this context. Additionally, our methodology integrates a bootstrap approach reminiscent of Analog bits~\cite{Analog_bits}, which significantly enhances our model's optimization and effectively narrows the gap between training and inference phases."

To evaluate our approach, we introduce the Inference Flow Consistency ($\ifc$) metric, reflecting \xflow severity. We also utilize Inception Score (IS)~\cite{IS} and Fréchet Inception Distance (FID)~\cite{FID} for assessing generation quality. Our models, trained from scratch and compared against baselines on \cifar~\cite{cifar10}, demonstrate not only an avoidance of \xflow but also an enhancement in generation quality. The contributions of this paper include:
\begin{itemize}
    \item We identify a widespread phenomenon in diffusion models, termed \xflow, leading to non-straight flow during inference that may generate OOD or suboptimal samples. 
    \item We attribute \xflow's origins to the instability of the target during the training process. Accordingly, we introduce the \ncd, a novel training and inference pipeline to mitigate the \xflow problem by enhancing input dimensionality.
    \vspace{3pt}
    \item Our experiments on both a toy model and the \cifar dataset demonstrate that our method not only improves the proposed $\ifc$ metric by addressing \xflow, but also significantly enhances other image evaluation metrics, such as IS and FID.
\end{itemize}

\section{Related work}

\subsection{Diffusion models}
Diffusion models, as generative models, learn the reverse denoising process from Gaussian noise to image distribution, achieved through either Markov~\cite{DDPM} or non-Markov operations~\cite{DDIM}. They are favored over other generative models like GAN~\cite{goodfellow2014generative} and VAE~\cite{vahdat2020nvae} due to their training stability and superior generation quality. Subsequent enhancements to these models primarily concern varied network architectures~\cite{karras2022elucidating}, noise schedulers or losses~\cite{iDDPM}, transition from image space diffusion to latent space~\cite{LDM}, and improved sampling techniques~\cite{lu2022dpm}, with little attention to the \xflow during training. Rectified Flow~\cite{rectified_flow} noted the mismatch in sampling across different inference steps, a significant distillation issue, but did not analyze it further. Instead, they proposed a workaround using a 2-rectified flow to fit another model to a non-crossing flow between source and target distributions, which depends on a well-trained diffusion model and requires additional retraining.
Our paper is the first to examine \xflow in diffusion and offer solutions. 

\subsection{Conditional Image Generation}
Conditioning techniques are instrumental in managing generated content~\cite{LDM,gao2023magicdrive,yang2022out,yang2022you}.
For diffusion models, score-based model~\cite{song2020score} proposes classifier guidance, which is an efficient method to balance controllability and fidelity using the gradients from a classifier, while classifier-free guidance~\cite{cfg}, being another important conditioning technique to diffusion models, trains both conditional and unconditional diffusion models, and combining their score to achieve better controllability.
ControlNet~\cite{ControlNet} employs pretrained encoding layers from billions of images as a backbone to learn diverse conditional controls, which is an architecture adopted in this paper.
Analog Bits~\cite{Analog_bits} introduces a technique that conditions the model on its own previously generated samples during iterative sampling, akin to our work.
However, Analog Bits mainly aims to enhance sample quality by reusing the previous target, while our focus is to introduce a condition in the training flow to prevent crossing issues. 
\begin{figure*}[]
  \centering
   \includegraphics[width=0.8\linewidth]{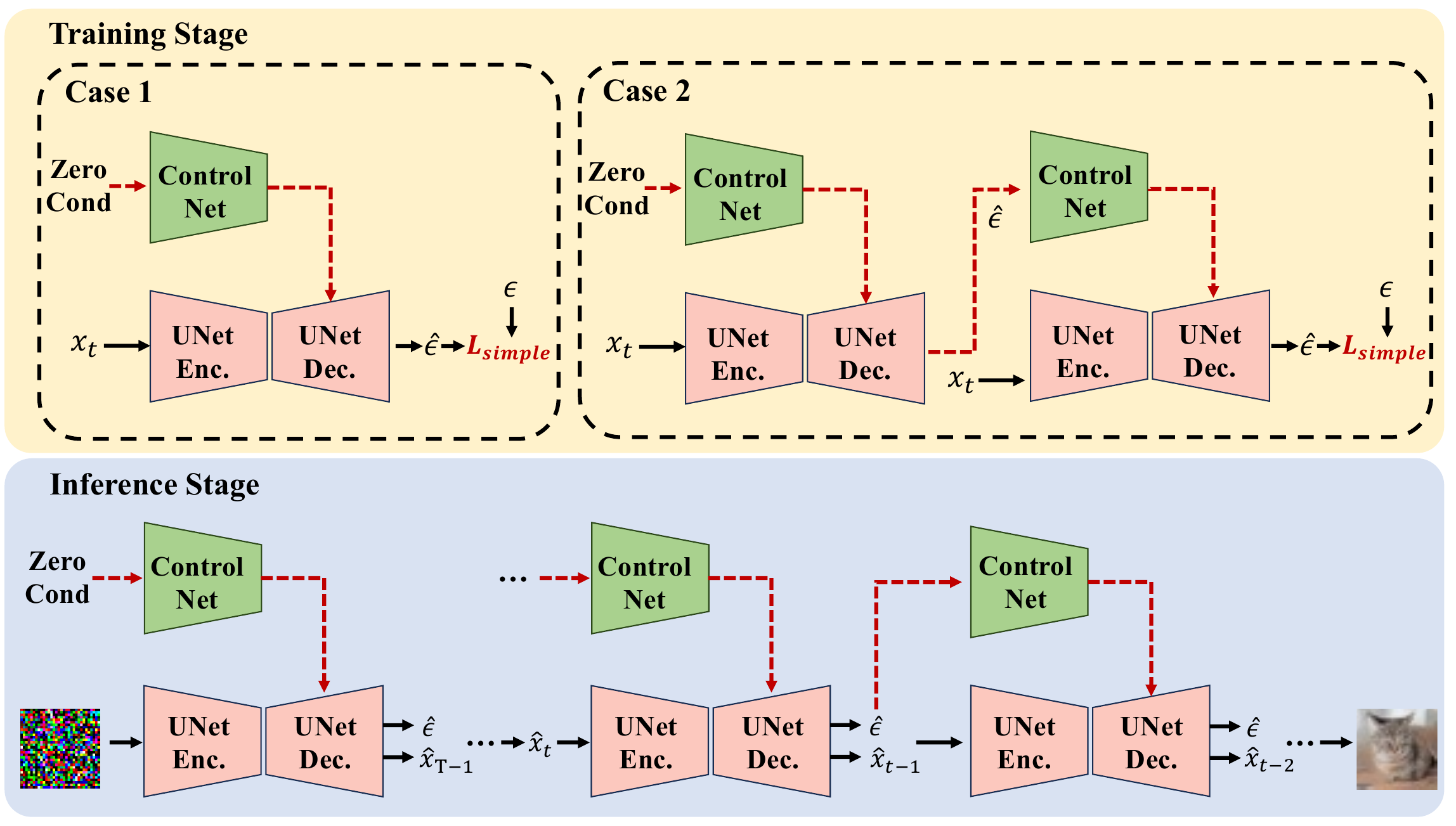}

   \vspace{-0.3cm}

   \caption{The overview of non-cross diffusion. \textbf{Training stage}: The training phase involves two cases. In Case 1, we utilize $\mathbf{0}$ as the condition and calculate loss function $L_{simple}$ as defined in Eq.~\ref{loss}. For Case 2, we first compute $\hat{\epsilon}$ using $\mathbf{0}$ as condition. Subsequently, $\hat{\epsilon}$ is employed as the condition to calculate $L_{simple}$. Throughout the training process, Case 1 is applied with a fixed probability $p$; otherwise, Case 2 is implemented.
   \textbf{Inference stage}: During the inference phase, $\textbf{0}$ is used as the condition in the initial denoising step. This is followed by iterative utilization of the estimated noise from the previous step as the condition for subsequent steps.}
   \label{overview}
\end{figure*}

\section{Method}
In this section, we start with a brief review of the formulation of DDPM~\cite{DDPM}. Next, we show the drawback of baseline flow and analyze the cause of \xflow.
Then, we introduce \ncd to avoid crossing by ascending dimension of input, together with training, inference, and network architecture of \ncd.
Finally, we introduce $\ifc$ for evaluating the consistency of the inference flow.

\subsection{Preliminary}
Given samples from data distribution $x_0\sim q(x_0)$, DDPM~\cite{DDPM} defines a forward noising process $q$ which produces latent variables ${x_1,\ldots,x_T}$ by gradually adding Gaussian noise with a variance schedule $\beta_t\in(0,1)$ as follows:
\vspace{-0.2cm}
\begin{gather}
    q(x_1,\ldots,x_T) \coloneqq \prod_{t=1}^T q(x_t|x_{t-1}), \\
    q(x_t|x_{t-1}) \coloneqq \mathcal{N}(x_t;\sqrt{1-\beta_t}x_{t-1},\beta_t \mathbf{I}).
    \label{eq_q}
\end{gather}
With $\alpha_t\coloneqq 1-\beta_t$ and $\Bar{\alpha}_t \coloneqq \prod_{s=0}^t \alpha_s$, the marginal $q(x_t|x_0)$ can be derived through Eq.~\ref{eq_q} as follows:
\begin{gather}
    q(x_t|x_0)= \mathcal{N}(x_t,\sqrt{\Bar{\alpha}_t}x_0,(1-\Bar{\alpha}_t)\mathbf{I}), \\
    x_t = \sqrt{\bar{\alpha}_t} x_0+\sqrt{1-\bar{\alpha}_t} \epsilon,
    \label{eq5}
\end{gather}
where $\epsilon \sim \mathcal{N}(\mathbf{0},\mathbf{I})$. Using Bayes theorem, we can calculate the posterior $q(x_{t-1}|x_t,x_0)$ in terms of $\beta_{t}, \alpha_{t}$ and $\bar{\alpha}_{t}$.
There are many different ways to parameterize $p_{\theta}$ to approximate the posterior, while DDPM~\cite{DDPM} chooses $p_{\theta}(x_{t-1}|x_t)=\mathcal{N}(x_{t-1};\boldsymbol{\mu}_{\theta}(x_t,t), \sigma_t^2\mathbf{I})$, and propose that predicting $\epsilon$ works best with a loss function:
\begin{gather}
    L_{simple} = E_{t,x_0,\epsilon}[\| \epsilon - \epsilon_\theta(x_t,t)\|^2],
\label{loss}
\end{gather}
where $\boldsymbol{\mu}_\theta(x_t,t) = \frac{1}{\sqrt{\alpha}_t}(x_t - \frac{\beta_t}{\sqrt{1-\bar{\alpha}_t}}\epsilon_\theta(x_t,t))$


\begin{figure*}[t]
  \centering
  \subfloat[]{
    \begin{minipage}[t]{0.24\linewidth}
    \centering
    \includegraphics[width=1\textwidth]{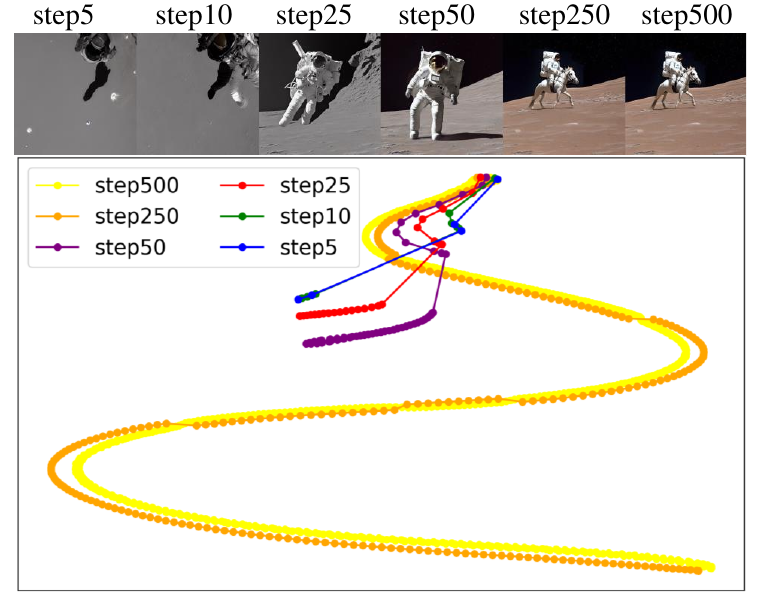}
    \end{minipage}
    \label{fig:xflow_sd_0}
}
    \subfloat[]{
    \begin{minipage}[t]{0.24\linewidth}
    \centering
    \includegraphics[width=1\textwidth]{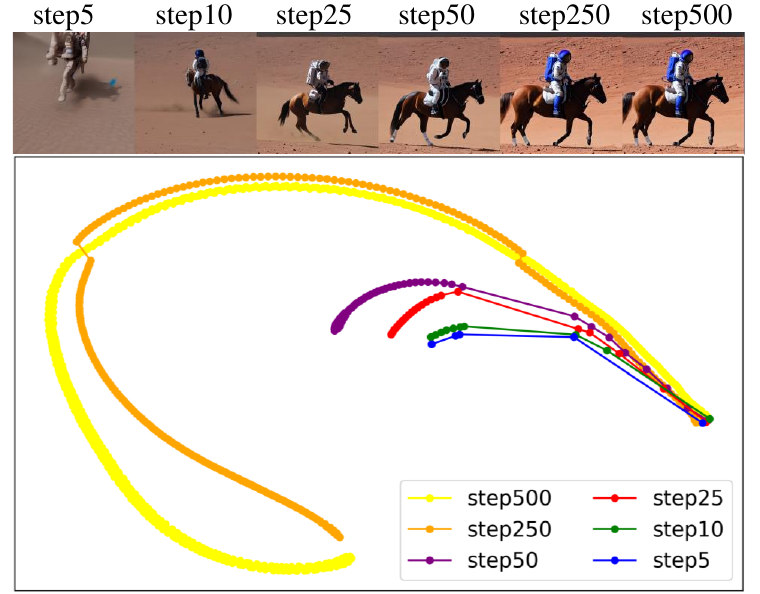}
    \end{minipage}
    \label{fig:xflow_sd_1}
}
    \subfloat[]{
    \begin{minipage}[t]{0.24\linewidth}
    \centering
    \includegraphics[width=1\textwidth]{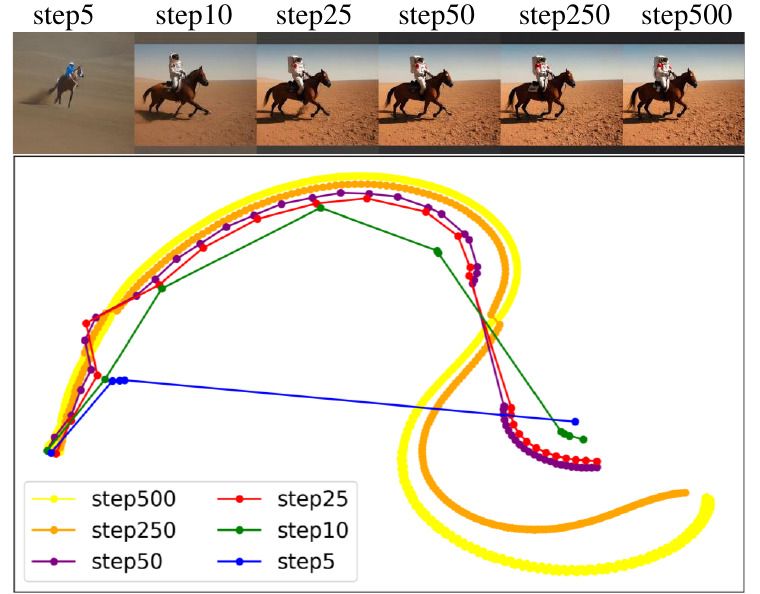}
    \end{minipage}
    \label{fig:xflow_sd_2}
}
    \subfloat[]{
    \begin{minipage}[t]{0.24\linewidth}
    \centering
    \includegraphics[width=1\textwidth]{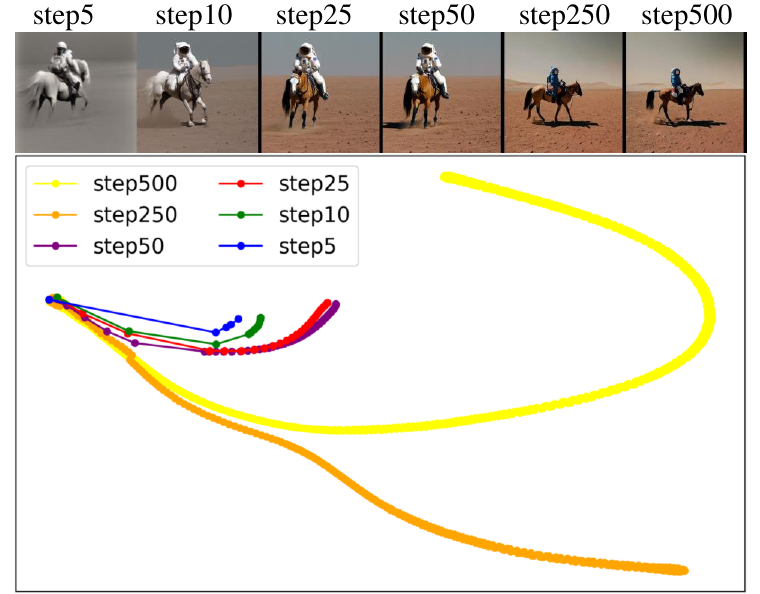}
    \end{minipage}
    \label{fig:xflow_sd_3}
}
  \caption{
    Generated images and inference flows of Stable Diffusion 1.5 using DDIM scheduler with prompt \textit{"a photo of an astronaut riding a horse on mars"} and negative prompt \textit{"bad, deformed, ugly, bad anotomy"}. (a)-(d) are generated with seed 0,1,2,3 respectively. The results demonstrate that the inference flows across different steps could be quite different at specific time $t$, which implies the influence of the \xflow.
  }
  \label{xflow_sd}
\end{figure*}

\subsection{Understanding Drawbacks of DDPM Flow}\label{sec:understanding}
\noindent\textbf{Training stage.} Given source distribution $\pi_0$ (\ie, $q(x_0)$) and target distribution $\pi_1$ (\ie, $\mathcal{N}(\mathbf{0}, \mathbf{I})$), we sample two data pairs $(x_0,x_T),(y_0,y_T) \sim \pi_0 \times \pi_1$. During the training stage, assume these two training flows cross at time step $t$ (\ie, $x_t = y_t$). Following Eq.~\ref{eq5}, we have:
\begin{gather}
    x_t = \sqrt{\bar{\alpha}_t} x_0+\sqrt{1-\bar{\alpha}_t} \epsilon_x, \\
    y_t = \sqrt{\bar{\alpha}_t} y_0+\sqrt{1-\bar{\alpha}_t} \epsilon_y.
\end{gather}
At the crossing point, both flows aim to minimize the loss function as follows during training:
\vspace{-0.2cm}
\begin{gather}
    L_{simple} = E_{t,x_0,\epsilon}[\| \epsilon - \epsilon_\theta(x_t,t)\|^2],
\end{gather}
For given $(x_0,x_T),(y_0,y_T)$ and cross point $t$, the loss can be reformulated as follows:
\begin{gather}
        L_1 = \frac{1}{2}\|\epsilon_x-\epsilon_\theta(x_t,t)\|^2+\frac{1}{2}\|\epsilon_y-\epsilon_\theta(y_t,t)\|^2,\label{equ:L1}
\end{gather}
where $\epsilon_x=x_T$ and $\epsilon_y=y_T$.
To check how the crossing point affects the optimization process, we simplify the target in the formulation of an optimization problem.
Since $\epsilon_\theta(x_t,t)=\epsilon_\theta(y_t,t)=\epsilon_\theta(z_t,t)$, we use the notion of $\epsilon_\theta$ to represent them all.
\begin{align}
    \theta^*&=\arg\min_\theta \frac{1}{2}\|\epsilon_x-\epsilon_\theta\|^2+\frac{1}{2}\|\epsilon_y-\epsilon_\theta\|^2 \\
            &=\arg\min_\theta \frac{1}{2}(\epsilon_x^2-2\epsilon_x\epsilon_\theta+\epsilon_\theta^2+\epsilon_y^2- 2\epsilon_y\epsilon_\theta+\epsilon_\theta^2) \\
            &=\arg\min_\theta \frac{1}{2}(\epsilon_x^2+\epsilon_y^2)-(\epsilon_x+\epsilon_y)\epsilon_\theta+\epsilon_\theta^2 \\
            &=\arg\min_\theta \frac{1}{4}(\epsilon_x^2+2\epsilon_x\epsilon_y+\epsilon_y^2)-(\epsilon_x+\epsilon_y)\epsilon_\theta+\epsilon_\theta^2 \\
            &=\arg\min_\theta \|\frac{\epsilon_x+\epsilon_y}{2}-\epsilon_\theta\|^2
            \label{wrong_target}
\end{align}
Hence, with the existence of a crossing point, the optimizing target is equivalent to $\frac{\epsilon_x+\epsilon_y}{2}$.
However, since $\epsilon_x\sim\mathcal{N}(\mathbf{0}, \mathbf{I})$ and $\epsilon_y\sim\mathcal{N}(\mathbf{0}, \mathbf{I})$, we have $\frac{\epsilon_x+\epsilon_y}{2} \sim \mathcal{N}(\mathbf{0}, \frac{\sqrt{2}}{2}\mathbf{I})$, which no longer follows standard normal distribution. This implies that at the crossing point, the model is given \textit{an incorrect target}, which will lead to \textit{ambiguity} in data generation (\ie, the denoising process).


\noindent\textbf{Inference stage.}
Considering the \textit{ambiguity} exists in a trained model, \xflow, where the generation flow may deviate from the correct direction, results in various failure cases, as illustrated in Fig.~\ref{fig:OOD}. We further visualize the inference flow of text-to-image Stable Diffusion in different steps. We save the intermediate latent of each step and visualize the inference flow by using T-SNE. As illustrated in Fig.~\ref{xflow_sd}, the visualization demonstrates that Stable Diffusion also shows various inference flows across different steps, which implies the influence of \xflow.
The results in Fig.~\ref{xflow_sd} demonstrate that the \xflow phenomenon occurs frequently within stable diffusion, which results in sub-optimal or even OOD synthesis.

Besides, we propose that the consequences of \xflow also depend on the timestep of inference.
Specifically, more inference steps, correlating with smaller strides, subtly affect inference flow because the deviation is also smaller, and subsequent steps can correct minor errors. 
On the contrary, fewer inference steps, leading to larger strides, significantly impact and alter the inference flow due to the crossing point, potentially generating inconsistent or OOD samples.
Such phenomena further decrease the determinism of diffusion models.


\subsection{Non-Cross Diffusion}\label{sec:method}
As analyzed in Sec.~\ref{sec:understanding}, \xflow is caused by incorrect training targets.
To solve \xflow, in this section, we introduce a new formulation of diffusion models that can avoid crossing points during training, namely \ncd.

Given the fact that latent variables are linear combinations of $x_{0}$ and $\epsilon$ as in Eq.~\ref{eq5}.
We can think of the issue with geometry, where training flows are line segments in 2D coordinates, as shown in Fig.~\ref{fig:OOD} (a), with the crossing point as the intersection of two segments.
From a basic geometrical concept, \ie, any two distinct lines in a plane can intersect at most once, as long as we can avoid the intersection once, the two segments will never intersect again.
Therefore, we aim to eliminate crossing points between any two different training flows, thereby maintaining the integrity and distinctiveness for all of them.

To operationalize this concept, we propose to ascend the dimension of model input.
The primary issue with Eq.~\ref{equ:L1} is that $\epsilon_\theta(x_t,t)=\epsilon_\theta(y_t,t)$ when $x_t=y_t$. To rectify this, we can introduce condition $c^x\neq c^y$ to ensure $\epsilon_\theta(x_t, c^x, t)\neq\epsilon_\theta(y_t, c^y, t)$ and prevent the training flow from crossing. The challenge is identifying $c^x\neq c^y$ given a cross point $t$ and training flow. To solve this, we use $x_i,y_i$ as conditions for each non-crossing step $i$ on the training flow, i.e., $c^x=x_i$ and $c^y=y_i$.
Specifically, given $x_{t}=y_{t}$, by sampling another point on the flow (\ie, $x_{i}$ and $y_{i}$), we have $x_{i} \neq y_{i}$, and thus $[x_i,x_t]\neq [y_i,y_t],\forall i\in[0,T]\setminus\{t\}$.
This reminds us that any other samples ($x_i$) from the same flow can be used for ascending dimensions.
This strategy effectively creates a multidimensional space where the likelihood of training flows intersecting is significantly reduced.

\noindent\textbf{Selection of Condition.}
$x_{i}$ is effective for ascending dimensions only if it is significantly different from $x_{t}$. Given the continuity of both linear combination and diffusion models, we propose to ascend the dimension with either initial noise $x_T$ (\ie, $\epsilon$) or the data point $x_0$.
Furthermore, we find the distance between randomly sampled noise is stable while the distance between data points may not.
Take image data as an example, for two randomly sampled noise $n_1,n_2\in \mathbb{R}^{H\times W\times C}$, we have $E[\|n_1-n_2\|^2] = 2CHW$.
Besides, we can only get the initial noise during the inference stage.
Therefore, using the initial noise $x_T$ for dimension ascending is more practical.


\subsection{Inference Flow Consistency}\label{sec:ifc}
To better evaluate the consistency of the inference flow for image generation, we propose a metric by computing the similarity between intermediate generated image $\hat{x}_0^t$ in timestep $t$ and the final generated image $\hat{x}_0$ based on peak signal-to-noise ratio (PSNR) as follows:
\begin{gather}
    \ifc = \frac{1}{T} \sum_{t=0}^{T} PSNR(\hat{x}_0^t,\hat{x}_0).
\end{gather}

\begin{figure*}[h]
  \centering
   \includegraphics[width=0.8\linewidth]{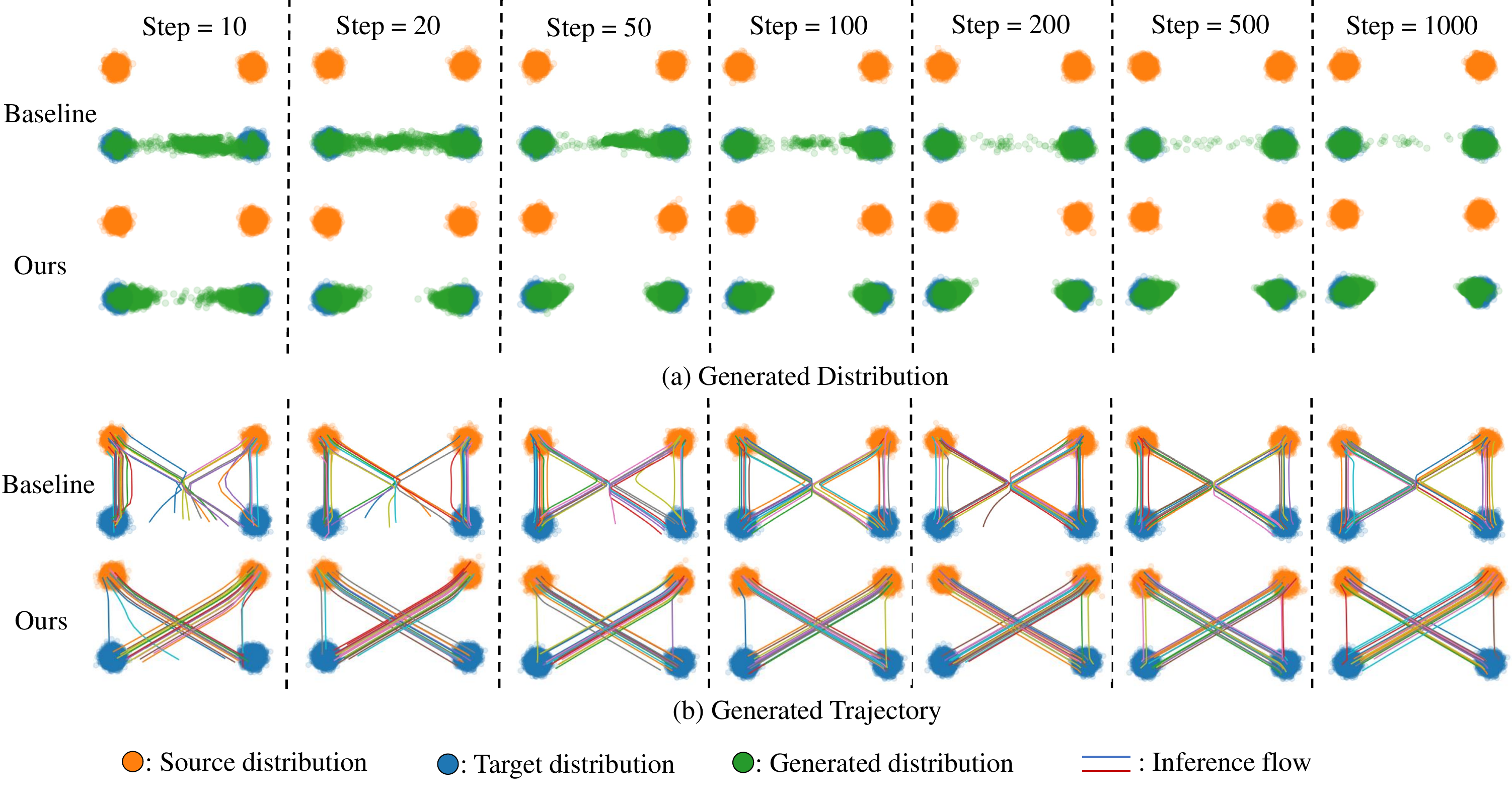}

   \vspace{-0.3cm}
   \caption{Results of the Toy Model. (a) Comparison of Generated Distributions: This panel illustrates the distributions generated by the baseline model and our proposed model. As the number of inference steps decreases, the baseline model tends to produce a significant number of out-of-distribution (OOD) samples. In contrast, our model effectively mitigates the generation of OOD samples. (b) Trajectory Analysis: This panel compares the generated trajectories of the baseline and our models. The baseline model's inference flow often redirects at the intersection point, leading to a target OOD distribution as the inference steps decrease. Our method, however, maintains a consistent direction in the inference model, thereby straightening the trajectory.}
   \label{toy_example}
\end{figure*}

\noindent\textbf{Training Stage.}
The cornerstone of our training strategy is to circumvent trivial solutions and avoid training collapse. To achieve this, we replace the use of initial noise $\epsilon$ with predicted noise $\hat{\epsilon}$. 
This substitution is critical in refining our model’s predictive accuracy since it can effectively avoid trivial solutions.
However, in the initial training phase, the substantial error $\|\epsilon-\hat{\epsilon}\|^2$ indicates $\hat{\epsilon}$'s poor estimation. Hence, a bootstrap strategy is introduced to $\hat{\epsilon}$ during training, preventing misleading estimation of $\hat{\epsilon}$ and thus enhancing learning robustness in the early stage.


As illustrated in Fig.~\ref{overview}, our training objective is formulated as follows:
\begin{gather}
    \min_\theta E_{t,x_t,\epsilon,}[\|\epsilon -\epsilon_\theta(x_t,\hat{\epsilon}_t,t)\|^2],
    \vspace{+0.1cm}
\end{gather}
 where $x_t = \sqrt{\bar{\alpha}_t} x_0+\sqrt{1-\bar{\alpha}_t} \epsilon$. 
During training, we apply the bootstrap as follows:
1) with a fixed probability $p$, we set $\hat{\epsilon}_t=\mathbf{0}$ (\ie, Case 1 in Fig.~\ref{overview});
2) at other cases, $\hat{\epsilon}_t$ is assigned the value of $\epsilon_\theta(x_t,\mathbf{0},t)$ (\ie, Case 2 in Fig.~\ref{overview}).
We do not back-propagate through estimated noise $\hat{\epsilon}_t$.


\noindent\textbf{Inference Stage.} 
As illustrated in Fig.~\ref{overview}, during the inference stage, to alleviate the computational costs, we use estimated noise in the previous step instead of the current step as the condition and iteratively predict $\hat{\epsilon}$ as follows:
\begin{gather}
    \hat{\epsilon}_{T} = \epsilon_\theta(\hat{x}_T,\mathbf{0},T), \\
    \hat{\epsilon}_t = \epsilon_\theta(\hat{x}_t,\hat{\epsilon}_{t+1},t), t<T.
\end{gather}
When the number of inference steps is large, the discrepancy between $\hat{\epsilon}_t$ and $\hat{\epsilon}_{t+1}$ is small, which ensures the performance of our method.

\noindent\textbf{Network Architecture.}
Inspired by ControlNet~\cite{ControlNet}, to efficiently use $\hat{\epsilon}_t$, \ncd employs an additive U-net branch, with $\hat{\epsilon}_t$ as input. For optimization, modifications are introduced, specifically removing all zero convolution layers and initializing the addictive encoder for $\hat{\epsilon}_t$ with the original U-net. The output is incorporated into the U-net decoder via addition. The whole network is trained end-to-end from scratch. 

A change in training flow direction at a specific timestep yields notable differences in pre- and post-change images, reducing PSNR. This can be effectively assessed for consistency across inference stages using our PSNR-based metric. 


\begin{table*}[]
\centering
\setlength{\tabcolsep}{10pt}
\scalebox{0.8}{
\begin{tabular}{l|ll|ll|ll|ll|ll|ll}
\hline
       & \multicolumn{2}{c|}{DDIM-1000} & \multicolumn{2}{c|}{DDIM-100} & \multicolumn{2}{c|}{DDIM-50}  & \multicolumn{2}{c|}{DDIM-20}  & \multicolumn{2}{c|}{DDIM-10}   & \multicolumn{2}{c}{DDIM-5}     \\ \hline
Method & IS             & FID           & IS            & FID           & IS            & FID           & IS            & FID           & IS            & FID            & IS            & FID            \\ \hline
iDDPM  & 9.02           & 4.70          & 8.99          & 5.65          & 8.89          & 6.61          & 8.59          & 9.82          & 8.20          & 15.91          & 7.09          & 31.37          \\
iDDPM$^\ddag$  & 9.10           & 4.82          & 8.93          & 5.75          & 8.79          & 6.71          & 8.65          & 9.89          & 8.14          & 16.06          & 7.08          & 31.21          \\
Ours   & \textbf{9.51}  & \textbf{2.88} & \textbf{9.22} & \textbf{3.93} & \textbf{9.10} & \textbf{5.31} & 8.77          & 9.87          & 7.97          & 20.63          & 6.20          & 50.25          \\
Ours$^\dag$   & 9.34           & 3.40          & 9.15          & 4.21          & 9.05          & 5.08          & \textbf{8.84} & \textbf{7.75} & \textbf{8.50} & \textbf{12.85} & \textbf{7.45} & \textbf{27.83} \\ \hline
\end{tabular}
}
\vspace{-0.1cm}
\caption{We compare the performance of baseline and our method. We generate 50k samples using DDIM with inference steps in \{1000, 100, 50, 20, 10, 5\}. $^\ddag$We expand the U-net encoder to ensure the same model size as ours. $^\dag$We use an inference strategy similar to the training stage. Specifically, we first give $\mathbf{0}$ as condition and get estimated noise $\hat{\epsilon_t}$, then we take $\hat{\epsilon_t}$ as condition and compute denoised image $\hat{x}_{t-1}$.}
\label{tab:FID}
\end{table*}

\section{Experiment}

In this section, we discuss our experimental results on toy examples (Sec.~\ref{sec:toy}) and image generation tasks (Sec.~\ref{sec:main exp}), as well as ablation studies (Sec.~\ref{sec:abl study}) and further discussion for \ncd (Sec.~\ref{sec:discussion}).

\subsection{Toy Examples}\label{sec:toy}
In this section, we follow the setting in Rectified Flow~\cite{rectified_flow}, drawing a training dataset from Gaussian mixture $\pi_0\times\pi_1$.
Given a sample $\{ x_0^i,x_1^i\}$ from $(X_0,X_1)\sim \pi_0\times\pi_1$, for baseline model, we train a 3-layer MLP $v_{\theta}(z,t)$ to transfer from $\pi_0$ to $\pi_1$ with l2-loss as follow:
\begin{gather*}
    \min_{\theta} \|v_{\theta}(x_t^i,t)-(x_1^i-x_0^i) \|_2, \\
    x_t^i = tx_1^i+(1-t)x_0^i, t\in [0,1).
\end{gather*}
Our method enhances this approach by incorporating an additional dimension with the estimated target as follows:
\begin{gather*}
    \min_{\theta} \|v_{\theta}([x_t^i,\hat{c}_t^i ],t)-(x_1^i-x_0^i) \|_2, \\
    \hat{c}_t^i = 
        \begin{cases}
        \mathbf{0}& \text{p$\leq$0.5}\\
        v_{\theta}([x_t^i,\mathbf{0}],t)& \text{otherwise}
        \end{cases}
\end{gather*}
with $p\sim U(0,1)$.
The inference process is also similar to our proposed method, \ie, we use the estimated target in the previous step as the condition.

\begin{figure*}
\centering
\subfloat[Baseline]{
\begin{minipage}[c]{0.45\linewidth}
    \centering
    \includegraphics[width=0.8\textwidth]{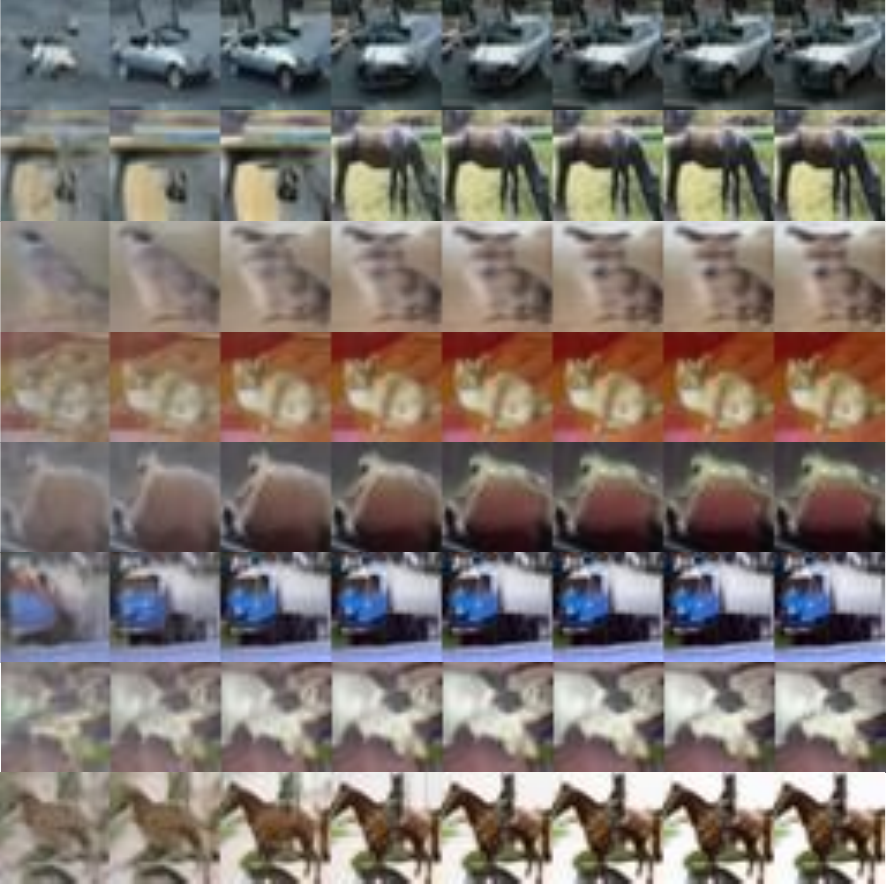}
\end{minipage}
\label{fig:flow-a}
}
\subfloat[Ours]{
\begin{minipage}[c]{0.45\linewidth}
    \centering
    \includegraphics[width=0.8\textwidth]{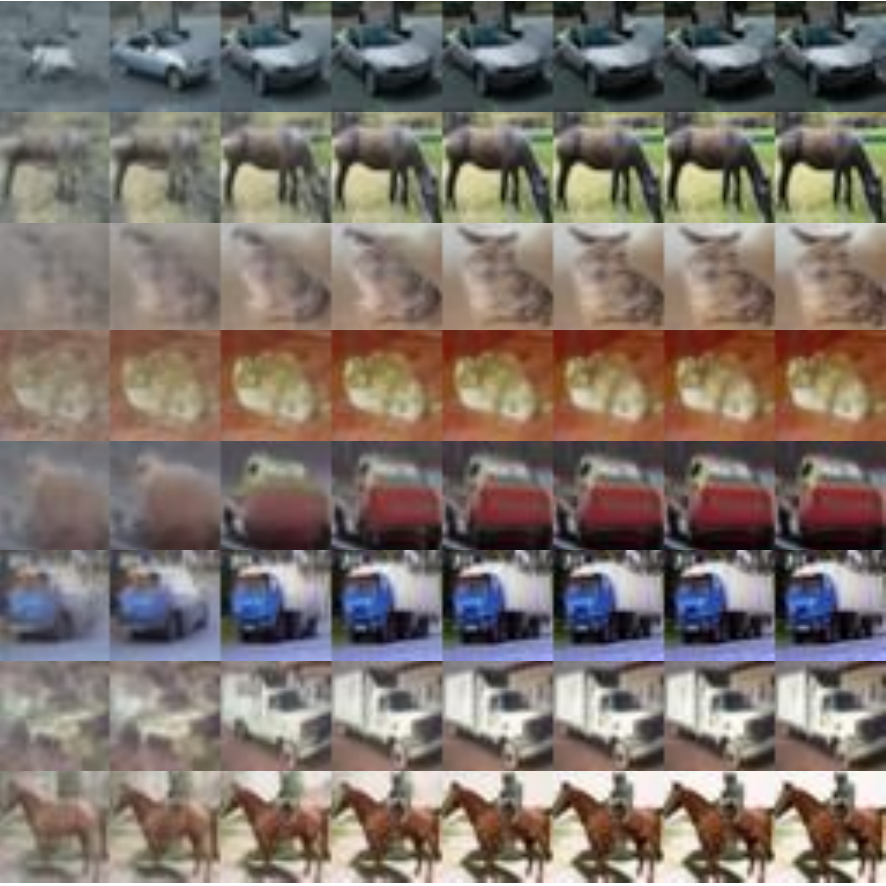}
\end{minipage}
\label{fig:flow-b}
}
  \vspace{-0.3cm}
  \caption{Here are the generated images using DDIM with inference steps in \{5, 10, 25, 50, 100, 200, 500, 1000\} on \cifar. For baseline method, the semantic information of image with small inference step and large inference step could be greatly different, which implies that the inference flow changes its direction at some timesteps.}
  \label{fig:cifar_flow}
\vspace{-0.3cm}
\end{figure*}

\begin{figure}[htb]
\centering
\subfloat[Baseline]{
\begin{minipage}[b]{0.45\linewidth}
    \centering
    \includegraphics[width=0.8\textwidth]{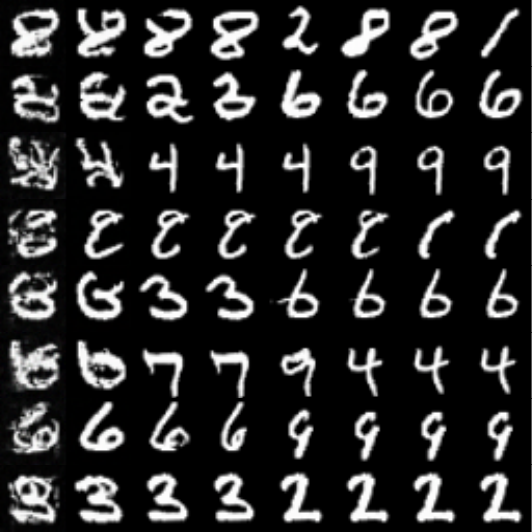}
\end{minipage}
\label{fig:mnist_flow-a}
}
\subfloat[Ours]{
\begin{minipage}[b]{0.45\linewidth}
    \centering
    \includegraphics[width=0.8\textwidth]{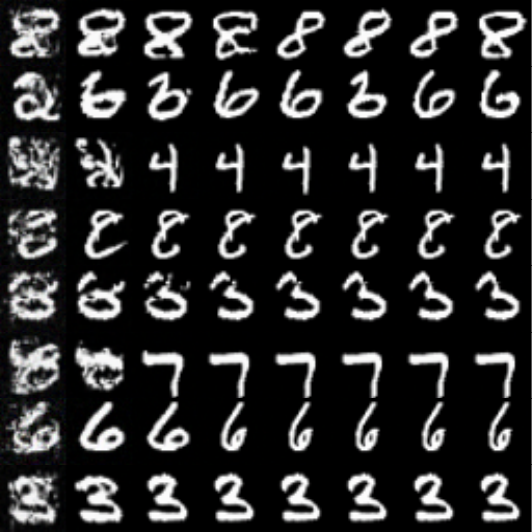}
\end{minipage}
\label{fig:mnist_flow-b}

}
  \caption{Here are the generated images using DDIM with inference steps in \{5, 10, 25, 50, 100, 200, 500, 1000\} on \mnist. 
  }
  \label{fig:mnist_flow}
\vspace{-0.2cm}
\end{figure}

\noindent\textbf{Results.}
As shown in Fig.~\ref{toy_example}, the baseline model's inference flow alters direction at the intersection of two flows due to an erroneous loss function (Eq.~\ref{wrong_target}), generating OOD samples. 
For toy examples, by adding an extra dimension using the estimated result, our method prevents training and inference flow intersection, maintaining consistent inference flow direction and effectively inhibiting OOD sample generation. 

\subsection{Experiments on Image Generation}\label{sec:main exp}
\noindent\textbf{Implementation Details.}
Our models are trained on \cifar~\cite{cifar10} and \mnist~\cite{mnist}, with \mnist images resized to $32\times32$. The fidelity of generated samples is evaluated using IS~\cite{IS} and FID~\cite{FID} and inference flow consistency with $\ifc$. As a baseline, we train iDDPM~\cite{iDDPM} from scratch with the same UNet. Training for \cifar follows iDDPM except using  $L_{simple}$ only and with 250k steps. For \mnist, training is similar to \cifar but with 100k steps. In this paper, we consider unconditional generations on each dataset (\ie w/o class labels).
\begin{table}[]
\centering
\setlength{\tabcolsep}{6pt}

\scalebox{0.8}{
\begin{tabular}{ll|ll|ll}
\hline
\multicolumn{2}{c|}{DDIM-1000}              & \multicolumn{2}{c|}{DDIM-100}               & \multicolumn{2}{c}{DDIM-50}                 \\ \hline
\multicolumn{1}{l|}{Method} & FID           & \multicolumn{1}{l|}{Method} & FID           & \multicolumn{1}{l|}{Method} & FID           \\ \hline
\multicolumn{1}{l|}{iDDPM}  & 8.02          & \multicolumn{1}{l|}{iDDPM}  & 8.26          & \multicolumn{1}{l|}{iDDPM}  & 9.12          \\
\multicolumn{1}{l|}{Ours}   & \textbf{7.13} & \multicolumn{1}{l|}{Ours}   & \textbf{7.72} & \multicolumn{1}{l|}{Ours}   & 9.06          \\
\multicolumn{1}{l|}{Ours$^\dag$}  & 7.52          & \multicolumn{1}{l|}{Ours$^\dag$}  & 7.91          & \multicolumn{1}{l|}{Ours$^\dag$}  & \textbf{8.76} \\ \hline
\end{tabular}
}
\vspace{-0.1cm}
\caption{We compare the performance of baseline and our method on \mnist. We generate 50k samples using DDIM with inference steps in \{1000, 100, 50\}. $^\dag$We use an inference strategy similar to the training stage.}
\label{tab:mnist_FID}
\vspace{-0.3cm}
\end{table}

\noindent\textbf{Comparison of Sampling Quality.}
Tables~\ref{tab:FID} and ~\ref{tab:mnist_FID} compare our model's performance with a baseline model in generating \cifar and \mnist images. Fig.~\ref{cifar_result} visualizes the generated \cifar images, where our model notably outperforms the baseline, especially at \{1000, 100, 50\} inference steps.
The decreased number of inference steps and increased strides enlarge the discrepancy in estimated noise between steps, introducing bias and performance decline during inference.
To counter this, we suggest conditioning on the current step's estimated noise. This augments image quality generated by our method even at smaller steps in \{20, 10, 5\}.
Besides, our model outperforms the baseline on \mnist at steps \{1000, 100, 50\}.
The adapted sampling strategy also enhances our model's performance on \mnist as the number of inference steps decreases.


\begin{figure*}
  \centering
  \subfloat[Baseline]{
    \begin{minipage}[t]{0.45\linewidth}
    \centering
    \includegraphics[width=0.8\textwidth]{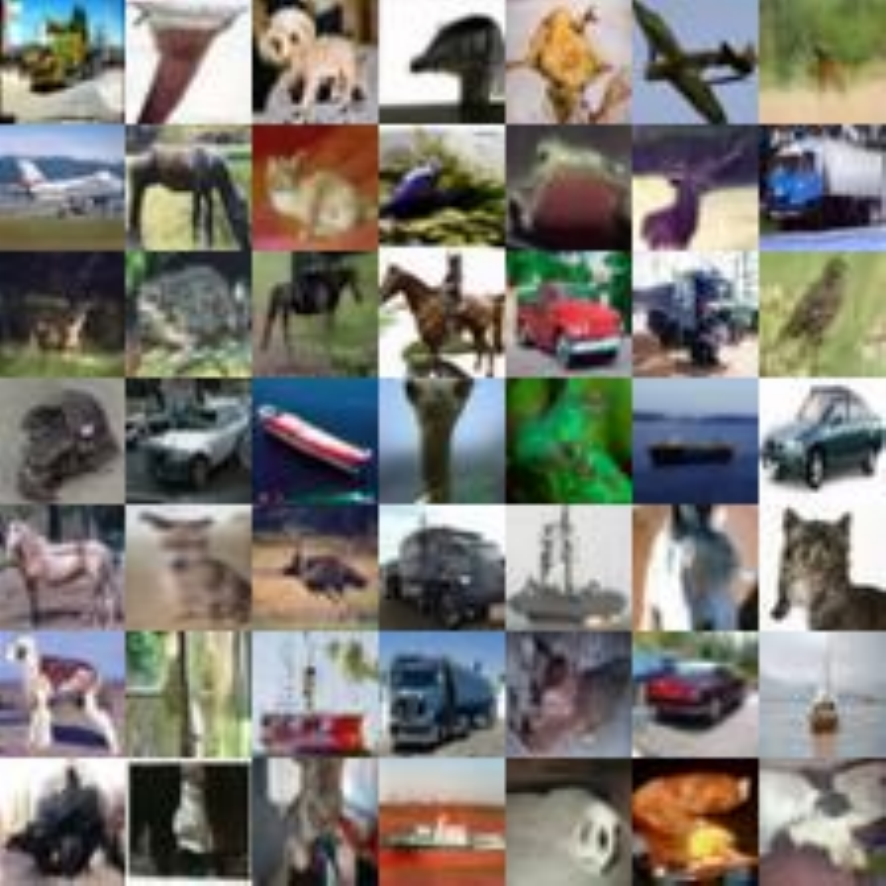}
    \end{minipage}
    \label{fig:result-a}
}
  \subfloat[Ours]{
    \begin{minipage}[t]{0.45\linewidth}
    \centering
    \includegraphics[width=0.8\textwidth]{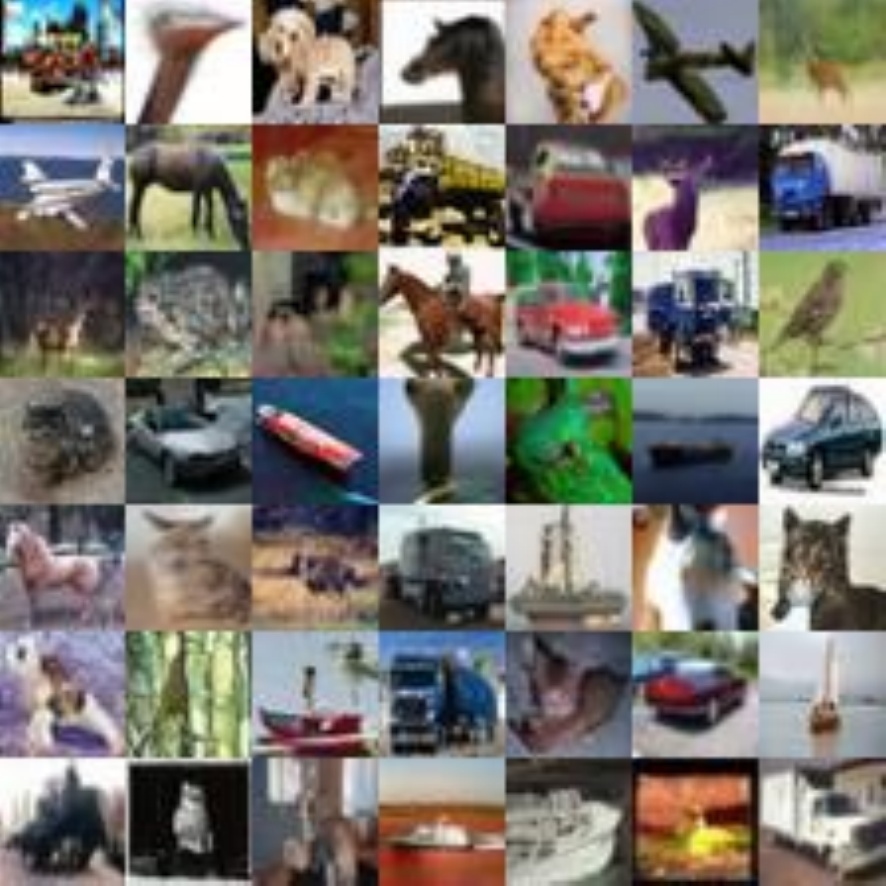}
    \end{minipage}
    \label{fig:result-b}
}
  
  \vspace{-0.35cm}
  \caption{Displayed are the generated images of baseline model and our model using DDIM with inference step of 1000 on \cifar. The results demonstrate our model's superior image generation capabilities, significantly reducing the occurrence of OOD samples.}
  \label{cifar_result}
\vspace{-0.4cm}
\end{figure*}

\noindent\textbf{Comparison of Inference Consistency.}
Table~\ref{tab:IFC} reveals our method's superior consistency over the baseline on \cifar at inference steps \{50, 20, 10, 5\}, in terms of $\ifc$, achieved by preventing \xflow. The impact of \xflow is minimal for larger steps (\{1000, 100\}), resulting in similar $\ifc$ between our method and the baseline.
Fig.~\ref{fig:cifar_flow} and Fig.~\ref{fig:mnist_flow} illustrate the visualization results of generated images under different steps on \cifar and \mnist. The results further demonstrate higher consistency compared with baselines, indicating a straighter inference flow.

\begin{table}[]
\centering
\scalebox{0.8}{
\begin{tabular}{lllllll}
\hline
\multicolumn{7}{c}{IFC}                                                                                                                                                            \\ \hline
\multicolumn{1}{l|}{DDIM Step} & \multicolumn{1}{c}{1000} & \multicolumn{1}{c}{100} & \multicolumn{1}{c}{50} & \multicolumn{1}{c}{20} & \multicolumn{1}{c}{10} & \multicolumn{1}{c}{5} \\ \hline
\multicolumn{1}{l|}{iDDPM}     & \textbf{28.58}           & 29.72                   & 30.85                  & 33.94                  & 39.51                  & 46.11                 \\
\multicolumn{1}{l|}{Ours}      & 28.37                    & \textbf{29.96}          & \textbf{31.4}          & \textbf{35.11}         & \textbf{40.21}         & \textbf{47.96}        \\ \hline
\end{tabular}
}
\caption{We compare the consistency of baseline and our method on \cifar. We generate 1000 samples using DDIM with inference steps in \{1000, 100, 50, 20, 10, 5\}. }
\label{tab:IFC}
\vspace{-0.4cm}
\end{table}


\subsection{Ablation Study}\label{sec:abl study}
In this section, we verify the effectiveness of each component through ablation studies.

\noindent\textbf{Ablation of bootstrap.}
Table~\ref{Ablation study} demonstrates the impact of the bootstrap strategy. We experimented with three settings: without bootstrap (w/o bootstrap, where $\hat{\epsilon}_t=\epsilon_\theta(x_t,\mathbf{0},t)$ is consistently applied), exponential schedule (exp-schedule, where the probability $p$ increases exponentially), and fixed probability (fix-probability, where $p=0.5$). The findings reveal that lacking a bootstrap strategy can notably degrade model performance. 

\begin{table}[t]
\centering 
\scalebox{0.9}{
\begin{tabular}{c|l|ll}
\hline
& Setting                & IS            & FID           \\ \hline
\multirow{3}{*}{Bootstrap}                                                     & w/o bootstrap          & 5.93          & 79.70         \\
& exp-schedule           & 9.22          & 5.24          \\
& fix-prob (Ours)         & \textbf{9.38} & \textbf{4.84} \\ \hline
\multirow{3}{*}{Condition}                                                     & $\hat{x}_0$ condition           & \textbf{9.64} & 20.21         \\
& mid. condition          & 7.15          & 59.09         \\
& $\hat{\epsilon}$ condition (Ours)    & 9.38          & \textbf{4.84} \\ \hline
\multirow{2}{*}{Network}                                                       & Double Unet            & \textbf{9.65} & 6.26          \\
& ControlNet-based (Ours) & 9.38          & \textbf{4.84} \\ \hline
\multirow{3}{*}{\begin{tabular}[c]{@{}c@{}}Inference \\ strategy\end{tabular}} & zero condition         & 9.06          & 6.43          \\
& ${\epsilon}$ condition          & 7.86          & 27.00         \\
& $\hat{\epsilon}$ condition (Ours)    & \textbf{9.38} & \textbf{4.84} \\ \hline
\end{tabular}
}
\caption{Ablation study. We generate 10k samples using DDIM with 1000 inference step.}
\label{Ablation study}
\vspace{-0.4cm}
\end{table}

\noindent\textbf{Ablation of condition.}
We scrutinize the impact of varying conditions on our method.
The $\hat{x}_0$ condition implies the model utilizes the predicted image as the condition, while the mid. condition uses the midpoint of training flow (\ie $\frac{\hat{x}_0+\hat{\epsilon}}{2}$). During inference, each technique employs its corresponding condition.
As in Table~\ref{Ablation study}, mid. performs suboptimally, presumably due to proximity to the training flows' intersection, as mentioned in Sec.~\ref{sec:method}.
Benefiting from its stability, $\hat{\epsilon}$ as a condition leads to a significant FID
improvement over $\hat{x}_0$ condition.

\noindent\textbf{Ablation of architecture.}
We also examine a Double U-net variant, \ie, doubling the U-net's input channel to accommodate $\hat{\epsilon}$ input. As Table~\ref{Ablation study} shows, the ControlNet-based model enhances FID by 1.42, likely because the two inputs serve distinct roles, and the additive branch enables effective differentiation between them, thereby facilitating training.  

\noindent\textbf{Ablation of inference strategy.}
We also consider several inference strategies applied after training (we utilize a trained $\hat{\epsilon}$-conditioned model). These inference strategies include the Zero condition (utilizing $\mathbf{0}$), the $\epsilon$ condition (using initial noise $\epsilon$), and the $\hat{\epsilon}$ condition (employing estimated noise $\hat{\epsilon}$).
Performance significantly deteriorates under the $\epsilon$ condition due to $\epsilon$-$\hat{\epsilon}$ discrepancy.
Though our model circumvents training ambiguity, the Zero condition marginally compromises performance during inference due to potential redirection at cross points in the inference flow. 

\subsection{Discussion}\label{sec:discussion}

This section offers an alternative perspective to understand our \ncd. The inclusion of $\hat{\epsilon}_{t}$ in the model input is seen as a strong conditional constraint, which we propose can reduce the variability of semantic information, thus lessening the severity of \xflow during inference. 

Specifically, the $\hat{\epsilon}_{t}$ condition in \ncd is highly specific at the pixel level, making it an exceptionally stringent constraint. Consequently, \ncd effectively mitigates the \xflow issue. An interesting question arises: how would \xflow be influenced by other forms of conditions with varying degrees of strength? To investigate this, we employed the pre-trained ControlNet model for empirical analysis and results.

Fig.~\ref{fig:OOD}(c) shows text conditional images from Stable Diffusion~\cite{LDM}, as well as pose conditional and depth conditional images from ControlNet~\cite{ControlNet}. The text condition images at steps \{5, 10, 25, 50, 250, 500\} present a significant shift in the semantic content of the images at each step.
In contrast, pose conditional images demonstrate more consistent semantic content across steps: the pose of super-hero remains the same but the background and style still show a large variation.
The depth conditional images keep the highest consistency across different steps, despite some variability in details such as the background and pattern.
Therefore, we hypothesize that stronger conditions ease the severity of \xflow.

\section{Conclusion}

In this study, we have addressed the \xflow phenomenon in diffusion models, characterized by deviations in generative flow that result in semantic inconsistencies and suboptimal image generation. Our novel approach, `\ncd', innovates in the realm of generative modeling by adopting ordinary differential equation models. 
Our empirical investigations, including both a toy example and the \cifar image dataset, demonstrate the substantial efficacy of the \ncd approach. The results show a marked reduction in semantic inconsistencies at various inference stages and significant improvements in the overall performance of diffusion models.

\noindent\textbf{Looking Ahead.}
The identification of \xflow as a critical issue during inference opens new avenues for research and application optimization. 
Despite the effectiveness of the proposed \ncd approach on mitigating \xflow, we acknowledge the challenges associated with retraining large-scale diffusion models such as Stable Diffusion. However, we are optimistic that future research will find ways to integrate these improvements into existing models, potentially circumventing the need for extensive retraining. This paper lays the groundwork for such advancements, aiming to enhance the reliability and quality of diffusion model outputs.

\iffalse
In conclusion, this paper identifies the \xflow phenomenon, \ie diffusion models often face issues related to deviations from a straight generative flow, leading to semantic inconsistencies and suboptimal generations.
To address this, we introduce `\ncd', a novel approach in generative modeling to learn ordinary differential equation models.
By promoting the dimension of input, \ncd effectively connects points sampled from two distributions without crossing paths.
To evaluate the severeness of \xflow, we propose the $\ifc$ metric based on the consistency between generations.
Empirical results on both toy example and image dataset \cifar prove the effectiveness of the \ncd approach, demonstrating a significant decrease in semantic inconsistencies at various inference stages and a substantial improvement in the overall performance of diffusion models.

\noindent\textbf{Future Work}
In this paper, we identify \xflow as a significant issue affecting many applications during inference. We analyze its root cause, propose an effective mitigation pipeline, and encourage further research for optimization such as dimension promotion and better model architecture. Although our solution requires retraining, which may be impractical for large-scale diffusion models like stable diffusion~\cite{LDM}, we hope future research can rectify this issue on pre-existing models to avoid retraining.
\fi

{\small
\bibliographystyle{ieee_fullname}
\bibliography{egbib}
}

\end{document}